\documentclass[a4paper]{article}

\usepackage[utf8]{inputenc}
\usepackage{INTERSPEECH2018}
\usepackage{multirow}
\usepackage[table]{xcolor}
\usepackage{collcell}
\usepackage{hhline}
\usepackage{pgf}
\usepackage{multirow}

\def\colorModel{hsb} %You can use rgb or hsb

\newcommand\ColCell[1]{
  \pgfmathparse{#1<49?1:0}  %Threshold for changing the font color into the cells
    \ifnum\pgfmathresult=0\relax\color{white}\fi
  \pgfmathsetmacro\compA{0}      %Component R or H
  \pgfmathsetmacro\compB{#1/100} %Component G or S
  \pgfmathsetmacro\compC{1}      %Component B or B
  \edef\x{\noexpand\centering\noexpand\cellcolor[\colorModel]{\compA,\compB,\compC}}\x #1
  } 
\newcolumntype{E}{>{\collectcell\ColCell}m{0.4cm}<{\endcollectcell}}  %Cell width
\newcommand*\rot{\rotatebox{0}}

\title{Punctuation Prediction in Spontaneous Conversations: \\ Can We Mitigate ASR Errors with Retrofitted Word Embeddings?}

\name{
Łukasz Augustyniak$^1$$^2$,
Piotr Szyma\'nski$^1$$^2$,
Mikołaj Morzy$^3$,
Piotr \.Zelasko$^4$, 
Adrian Szymczak$^1$,
Jan~Mizgajski$^1$$^3$, 
Yishay Carmiel$^1$,
Najim~Dehak$^4$
}
\address{
  $^1$ Avaya, USA \\
  $^2$ Wrocław University of Technology, Department of Computational Intelligence, Wrocław, Poland \\
  $^3$ Poznan University of Technology, Poznan, Poland \\
  $^4$ Center for Language and Speech Processing, The Johns Hopkins University, Baltimore, MD, USA
}

\email{\{lukasz.augustyniak,piotr.szymanski\}@pwr.edu.pl}

\begin{document}

\maketitle

\begin{abstract}
\textbf{Automatic Speech Recognition (ASR) systems introduce word errors, which often confuse punctuation prediction models, turning punctuation restoration into a~challenging task. These errors usually take the form of homonyms. We show how retrofitting of the word embeddings on the domain-specific data can mitigate ASR errors. Our main contribution is a~method for better alignment of homonym embeddings and the validation of the presented method on the punctuation prediction task. We record the absolute improvement in punctuation prediction accuracy between 6.2\% (for question marks) to 9\% (for periods) when compared with the state-of-the-art model.}
\end{abstract}


\section{Introduction}
\label{sec:introduction}

Automatic Speech Recognition (ASR) systems are becoming ubiquitous not only in the human-computer interaction systems, such as voice assistants or dictation tools, but also in systems processing human-human conversations. The abundance of the available audio data makes it very tempting to use conversation transcripts as input data for spoken language understanding. Most commercially available ASR systems do not produce any punctuation or capitalization of output transcripts, which is a~serious limitation with respect to many downstream tasks. The pre-requisite for true spoken language understanding is the ability to comprehend spoken utterances, both at the semantic and syntactic level. The latter requires a~robust dependency parsing, which, in turn, partially relies on correct punctuation. 

Punctuation is also indispensable for intent annotation. Consider the task of annotating instances of a~negative sentiment in call transcripts. The phrase "\emph{nobody came back and I~don't like I~said, he didn't leave a~slip}" would be incorrectly marked as an instance of negative sentiment due to the presence of the utterance "\emph{I~don't like}," whereas in reality the phrase should be punctuated as "\emph{nobody came back and I~don't, like I~said, he didn't leave a~slip}." In general, missing punctuation can introduce errors for phrases with personal references ("\emph{let's eat grandma}" vs. "\emph{let's eat, grandma}"), enumerations ("\emph{I~love cooking my family and pets}" vs. "\emph{I~love cooking, my family, and pets}"), or prepositions at sentence boundaries ("\emph{taken care of the refund}" vs. "\emph{taken care of. The refund}"). The problem of the inherent lack of punctuation is exacerbated by the presence of stochastic ASR errors, which for spontaneous human-human conversations can amount to 15\%-20\% of transcribed words. Consider the following transcript: "\emph{Hi my name is e~agent will do I~have pleasure speaking with today}". It may be very challenging to correctly introduce punctuation since the actual utterance is "\emph{Hi my name is Adrian who do I~have pleasure speaking with today}."

Spontaneous speech is very different not only from the written language, but from other types of speech as well~\cite{Shriberg2005}. Scripted speech and human-computer conversations tend to have well-defined structure with clear demarcation of sentence-like units, correct SVO (subject-verb-object) structures, and limited vocabulary. Spontaneous speech, on the other hand, is filled with all types of disfluencies which can account for 5\% of all words and affect more than 30\% of utterances. These disfluencies, which include backchannel markers, coordinating conjunctions, discourse markers, or filled pauses, hinder transcript translation, summarization, information extraction, or readability of transcripts. At the same time, the disfluencies are known to play an important role in the management of interactions, for instance, in upholding the turn by a~speaker. One should also not forget that the rate of disfluency production varies significantly between people~\cite{igras2016structure}, making it difficult to introduce a~general model of disfluencies.

An important, yet often overlooked, aspect of spontaneous human-human conversations is the overlap of utterances~\cite{Shriberg2001}. As previous research suggests, the overlap in conversational speech is substantial, and it is as frequent in phone conversations between strangers as in face-to-face meetings between close acquaintances. The presence of the overlap makes diarization of speech more difficult, which also affects the ability to model the turn-taking realistically. Interestingly, we can see that the occurrence of the overlap changes the structure of utterances as the speakers react dynamically to the interruptions by repeating certain phrases, correcting, or deleting them. An example of such a~change might be the increased number of question marks -- speakers use questions much more often than in regular speech, not only due to the conversational nature of the exchange, but also the need to request acknowledgements of comprehension or verify comprehension by paraphrasing.

In principle, any punctuated text can be used to train a~punctuation model. Unfortunately, most of the available textual corpora~are not representative of spontaneous speech. Patterns learned from Wikipedia, Web Crawl, or news corpora, hardly generalize to the transcripts of spontaneous conversations. Obtaining new annotated datasets is also very challenging. Raw conversational transcripts are illegible and manual restoration of punctuation marks is both time-consuming and expensive. 

The main hypothesis underlying our approach is that it is possible to mitigate stochastic ASR errors by retrofitting static word embeddings to the domain of the application. During punctuation prediction we cannot correct the ASR errors, but the retrofitted representation of words allows us to improve the accuracy of punctuation prediction models. As our main contribution, we validate this hypothesis by showing how pre-trained GloVe embeddings can be fine-tuned to the domain of call center conversations by using Mittens~\cite{mittens}, and how this retrofitting improves the accuracy of punctuation prediction in transcribed calls. We compare our approach with two state-of-the-art solutions (a~bi-directional LSTM model and a~CNN with pre-trained embeddings), showing significant improvements in punctuation prediction accuracy.

\section{Related work}
\label{sec:related.work}

The simplest form of punctuation prediction is the discovery of sentence-like unit boundaries, where the problem is the binary classification (with "period" and "space" classes). Historically, many different techniques have been tried, for instance, Word Confusion Matrices~\cite{hillard2004improving}, Maximum Entropy Models~\cite{huang2002maximum}, Conditional Random Fields~\cite{lu2010better}, Hidden Markov Models~\cite{yamron1998hidden}, and mixtures of probability models~\cite{liu2004comparing}. Features used to detect sentence boundaries included both linguistic information (n-gram language models, turn markers, part of speech annotations)~\cite{stolcke1996automatic} and prosodic features~\cite{christensen2001punctuation, shriberg2000prosody, wang2004multi}.

The advent of modern deep neural networks introduced unprecedented advancements in punctuation prediction. Recurrent neural networks quickly surpassed previous state-of-the-art models. Apart from incorporating word embeddings into punctuation prediction~\cite{che2016punctuation}, these models initially employed LSTM architectures to predict punctuation marks using longer contexts~\cite{tilk2015lstm, tilk2016bidirectional}. More recently, simpler architectures have proven to be sufficiently robust. Character-level convolutional neural networks (CNNs) can restore punctuation marks efficiently and these models do not suffer from out-of-vocabulary tokens or long inference times. At the same time, CNNs struggle to catch longer contexts necessary to restore question marks~\cite{gale2017experiments}. In the last two years, transformer-based models have been gaining popularity for both general punctuation prediction~\cite{nguyen2019fast} and more specific tasks, such as question prediction~\cite{cai2019question} or disfluency removal~\cite{chen2020controllable}. The popularity of transformer-based models can be easily explained by the usefulness of the mechanism of attention in punctuation prediction~\cite{tilk2016bidirectional,oktem2017attentional}. Another area~of active research is the reframing of the punctuation prediction task in terms of machine translation. These works are mainly driven by real-time translation services~\cite{cho2015punctuation, cho2017nmt, peitz2011modeling}.

We should stress that most of the previous works focused on punctuation prediction for speech, not for conversations, which makes the results incomparable with our case. Models are usually trained on audio recordings with already available high-quality transcriptions -- these are often examples of scripted speech, for instance, TED talk transcripts or transcripts of speeches in the European Parliament. Golden standard conversational transcripts are, as we have already mentioned, very expensive to produce and only a~handful of such datasets exists. 

\section{Methods}
\label{sec:methods}

\subsection{Data}
\label{subsec:data}

Our primary training data~is the Fisher corpus~\cite{cieri2004fisher} due to its adequately punctuated transcripts. The Fisher corpus creation protocol relied upon a~vast number of participants, each making a~few short calls. Typically, the speakers would not know each other personally, which maximized inter-speaker variation and vocabulary breadth, although it also increased the formality of speech. The goal was to provide a~representative distribution of subjects across a~variety of demographic categories, including gender, age, dialect region, and English fluency. Punctuation classes in the Fisher corpus are highly unbalanced (see Table~\ref{tab:punctuationClasses}), which is typical for conversational speech. Hence, the Fisher corpus is a~proper training and evaluation dataset for the punctuation prediction in the ASR transcripts. 

\begin{table}[th]
    \caption{The distribution of punctuation classes in the Fisher corpus.}
    \label{tab:punctuationClasses}
    \centering
    \begin{tabular}{ c r c }
        \toprule
        \textbf{Class} & \textbf{Count} & \textbf{Percentage} \\
        \midrule
        $\epsilon$ (blank)& 1 429 905 & 79.1\% \\
        ,             & 208 289  & 11.5\% \\
        .             & 148 624  & 8.2\%  \\
        ?             & 22 182   & 1.2\%  \\
        \bottomrule
    \end{tabular}
\end{table}

To fit the Fisher corpus into our model definition, we need to combine information from the time-annotated and punctuated transcripts. The first step is computing the forced alignment of the time-annotated transcripts to obtain word-level information about starting times and durations of words. For that purpose, we use the Kaldi ASR toolkit~\cite{povey2011kaldi} with an LSTM-TDNN acoustic model trained with lattice-free Maximum Mutual Information (MMI) criterion~\cite{povey2016purely}. In order to minimize the differences between the two transcript versions, we edited the Fisher corpus preparation script not to exclude single-word utterances and the text in parentheses. We retained blanks (no punctuation), periods, commas, and question marks. Other punctuation classes (e.g., exclamation marks or ellipses) were converted to blanks due to their low frequency. Finally, we aligned the time-annotated transcript with the punctuated transcript using the Needleman-Wunsch algorithm~\cite{needleman1970general}. 

\subsection{Features}
\label{subsec:features}

We represent the conversation $\mathcal{C}$, as a~chronologically ordered sequence $\mathcal{C} = \left[w_1,w_2,\ldots,w_n\right]$ of words $w_i=\langle t_i, c_i, s_i, d_i, p_i \rangle$, where: $t_i$ is the textual representation of the word $w_i$; the binary feature $c_i \in \{A,B\}$ represents the conversation side uttering the word $w_i$; the real number $s_i$ represents the time offset at which the word $w_i$ started; the real number $d_i$ represents the duration of the word $w_i$; and $p_i$ is the punctuation mark which appears after the word $w_i$. The use of mixed conversation sides yields to efficient representation of interjections, interruptions, overlap, and simultaneous speech. The punctuation marks are only known at the training time and are being predicted during inference. We treat the punctuation prediction problem as a~sequence tagging task.

We use three types of features in our models. First and foremost, we use static word embeddings. We choose 300-dimensional pre-trained GloVe~\cite{pennington2014glove} vectors which are being retrofitted (see Section~\ref{sec:methods}). Next, we use the interval (offset) between the start of the current word and the end of the previous word, and the duration of the present word. We standardize both of these features w.r.t. other words uttered by the same speaker in the same conversation. As a~result, the pauses are not modelled explicitly as word tokens, but they are inferred by the model based on word timings (offsets and durations). 

\subsection{GloVe retrofitting with Mittens}
\label{sec:mittens}

Word embeddings have become a widely used transfer learning approach for language processing. While different strategies for training the embeddings exist, the general premise is to encode the probability that a~word will occur in the context of other words using dense vectors. These probabilities are estimated within large scale corpora, such as Wikipedia text, product reviews, or social media. In this work, we use pre-trained GloVe embeddings trained on the Common Crawl dataset consisting of 2.6 billion textual documents scraped from the Web. This general written text is a~great resource to capture large-scale relations between words and their contexts. It is problematic, however, that the general language model trained on such corpus is not well aligned with the domain-specific tasks.

In order to tackle the problem of language models in health-care data, Dingwall and Potts \cite{mittens} devised a~retrofitting model for static embeddings. The authors reformulate the task of training GloVe embeddings in a~matrix form and they propose the extension of the objective function to take into account domain-specific data. In their formulation, the embedding $\widehat{w}_{i}$ of the word $w_i$ is the sum of the original vector for the case when the word is predicted in the context and the $\widetilde{w}_{i}$ embedding of the word $w_i$ if the word is the part of the context.  As a~result, they propose to retrofit the original embeddings by adding the square distance penalty against the new vector $\widehat{v_i}$ for all the words $v_i \in V$, where $V$ is the vocabulary of the domain-specific corpus and $J$ is the matrix form of existing embeddings. The weight $\mu$ can be used to control the retrofitting impact. The new objective function $J_{\text{Mittens}}$ becomes:

\begin{equation*}
  J_{\text{Mittens}} =
  J + \mu \sum_{v_i \in V}
         \|\widehat{w}_{i} - \widehat{v_i}\|^{2}.
  \label{eq:mittens}
\end{equation*}

In our case, the punctuation in conversational transcripts is substantially different from the punctuation found in the Common Crawl corpus. Additionally, transcripts suffer from ASR errors, where certain homonym sequences can be put in place of the correct transcription. However, if these homonym sequences are short (1-2 words), we hypothesize that they can be corrected using the retrofitting, as the ASR errors should happen in the same contexts as correct words. If our hypothesis is correct, it will allow us to overcome the problem of the absence of quality embeddings trained on the domain-specific corpora.

\section{Experiments}
\label{sec:experiments}

\subsection{Models}
\label{subsubsec:models}

All our models share the same convolutional neural network (CNN) architecture. Each model uses several layers of 1D convolutions which can be interpreted as fully-connected layers processing the input in small windows. Each layer is followed by a~SELU activation~\cite{klambauer2017self}, which yielded a~small improvement over batch normalization~\cite{ioffe2015batch} with ReLU~\cite{nair2010rectified}. We have experimented with many combinations of different numbers of layers, filter sizes, and other hyper-parameters. The best and most stable results have been achieved with six 1D convolutional layers with the filter size of 128 and zero padding. We use kernel sizes equal to 3 for all layers but the last one, where the kernel size is equal to 20. Note that these kernels could mix words from both sides of the conversation. All hidden layers use the dilation rate equal to 2. We have also experimented with model regularization. Firstly, we have added 0.5 dropout before the softmax layer. Secondly, we have used the weight decay (l2 with 0.001 weight) for the softmax layer. We have added the Gaussian noise with $\sigma=0.1$ before the last softmax activation and SELU activations. Finally, we have used SELU activations that constrain the weights to a~$N(0,1)$ distribution. The final layer of our model is the fully-connected layer with softmax activation. It is applied separately at each time step to retrieve punctuation prediction for a~given word.

\subsection{Training}
\label{subsec:training}

The models are implemented in Keras~\cite{chollet2015keras} with Tensorflow~\cite{abadi2016tensorflow} back-end. During the training, the weights are updated using the Adam optimizer \cite{kingma2014adam}. We use categorical cross-entropy as the loss function. We reduce (by the factor of 0.5) the learning rate (the minimum learning rate is set to $1e-5$) with the patience set to 3 epochs. We use the batch size of 256, and each sample is 200 words long. The Fisher corpus is divided into training, validation, and test sets in proportions 80:10:10. 

\section{Results}\label{sec:results}

We compare three variants of CNN-based models with CNN and BiLSTM baselines~\cite{Zelasko2018}. Each model is evaluated using precision, recall, and F1 score for each punctuation class separately. The results are presented in Table~\ref{tab:results}. In addition, we investigate the similarity of words when represented using the original GloVe vectors vs. vectors retrofitted using Mittens. 

\subsection{Model comparison}

\begin{table}[h]
    \caption{
        The per-class precision, recall, and F1-score (in \%) achieved by the compared models on the pre-trained GloVe embeddings with additional retrofitting using Mittens.}
    \label{tab:results}
    \centering
    \begin{tabular}{ r r r r r }
        \toprule
        \textbf{Model} & \textbf{Class} & \textbf{Precision} & \textbf{Recall} & \textbf{F1} \\
        \midrule
        \multirow{4}{*}{CNN-baseline}
        & $\epsilon$ & 92.7 & \textbf{95.8} & 94.2 \\
        & . & 65.5 & 58.7 & 61.9 \\
        & ? & 67.5 & 49.0 & 56.8 \\
        & , & 66.6 & 55.1 & 60.3 \\
        \midrule
        \multirow{4}{*}{BiLSTM-baseline}
        & $\epsilon$ & \textbf{\textit{93.5}} & 94.7 & 94.1 \\
        & . & 67.9 & 66.7 & 67.3 \\
        & ? & 64.7 & 54.6 & 59.2 \\
        & , & 68.2 & \textbf{64.1} & \textbf{66.1} \\
        \midrule
        \multirow{4}{*}{CNN-50k}
        & $\epsilon$ & 92.6 & 95.5 & 94.0 \\
        & . & 70.2 & 65.0 & 67.5 \\
        & ? & 70.2 & 51.7 & 59.5 \\
        & , & 69.7 & 60.8 & 65.0 \\
        \midrule
        \multirow{4}{*}{CNN-50k-mittens}
        & $\epsilon$ & 93.3 & 95.3 & 94.3 \\
        & . & 70.7 & \textbf{68.7} & 69.7 \\
        & ? & \textbf{72.8} & 53.7 & 61.8 \\
        & , & 69.3 & 62.4 & 65.7 \\
        \midrule
        \multirow{4}{*}{CNN-100k-mittens}
        & $\epsilon$ & 93.1 & 95.6 & \textbf{94.3} \\
        & . & \textbf{71.8} & 67.7 & \textbf{69.7} \\
        & ? & 71.2 & \textbf{55.2} & \textbf{62.2} \\
        & , & \textbf{69.9} & 61.9 & 65.7 \\
        \bottomrule
    \end{tabular}
\end{table}

As our baseline we choose the convolutional neural network and standard pre-trained GloVe embeddings for $50\,000$ most frequent words in GloVe training data. This is a~strong baseline which has proven to be a~viable solution for a~production-ready, real-time system~\cite{Zelasko2018}. Next, we constrain the selection of the $50\,000$ words to those which appear in the ASR vocabulary (CNN-50k model). This model language covers words that appear in the ASR vocabulary twice as frequently as the baseline model, and it achieves approx. 60\% coverage of the vocabulary. We observe the improvement of both precision and recall across all punctuation classes.

As the main experiment, we investigate the influence of the additional retrofitting of word embeddings using the domain-specific data (in our case, call center conversation transcripts). Our hypothesis is that the additional retrofitting puts erroneous pairs of homonyms (words and their corresponding ASR errors) closer in the embedding space. An example of an erroneous pair of homonyms are the words "\textit{cancer}" and "\textit{cancel}," for example, "\textit{I~have to cancer that appointment}" vs. "\textit{I~have to cancel that appointment}." 

After observing the general improvement in all metrics of the punctuation prediction, we have further extended the embeddings to cover almost the entire ASR vocabulary (approximately 20\% of the ASR vocabulary is missing from GloVe). The final model outperforms the baseline with respect to all three considered metrics while not imposing significantly increased resource requirements or incurring computational costs. 

Finally, we compare our CNN-based model with the BiLSTM model based on the GloVe embeddings. Our model outperforms this baseline with regards to precision and recall for all punctuation classes, except full stops. The additional advantage of using a~CNN-based model in the production environment is a~more straightforward parallelization compared to the BiLSTM architecture.

\subsection{GloVe similarity analysis}

To further validate the hypothesis that the retrofitting of word embeddings on the domain-specific data attenuates for the ASR errors by moving homonyms closer in the embedded space, we compute the cosine similarity between pairs of homonyms in the original and the retrofitted space. The results are presented in Table~\ref{tab:similarity}. As can be seen, the vectors are much closer in the retrofitted space than in the original GloVe space.

\begin{table}[h]
    \caption{Similarity between homonyms in the original GloVe space and in the retrofitted space.}
    \label{tab:similarity}
    \centering
    \begin{tabular}{rcc}
        \toprule
        \multirow{2}{*}{Homonyms} & \multicolumn{2}{c}{Cosine similarity}            \\
                                                       & original             & retrofitted \\
        \midrule
        I~have to cancel                               & \multirow{2}{*}{0.74} & \multirow{2}{*}{0.91}     \\
        I~have to cancer                               &               &                   \\
        \midrule
        cancel                                         & \multirow{2}{*}{0.1} & \multirow{2}{*}{0.18}     \\
        cancer                                         &               &                   \\
        \midrule
        by                                             & \multirow{2}{*}{0.16} & \multirow{2}{*}{0.4}     \\
        buy                                            &               &                    \\
        \midrule
        go to the court                                & \multirow{2}{*}{0.69} & \multirow{2}{*}{0.78}     \\
        got the card                                   &              &                  \\
        \midrule
        thank you                                      & \multirow{2}{*}{0.84} & \multirow{2}{*}{0.9}     \\
        think you                                      &               &  \\                
        \bottomrule
    \end{tabular}
\end{table}

\subsection{Confusion matrix analysis}

Table~\ref{table:confusion_matrix} presents the comparison of the confusion matrices for the CNN-baseline and the CNN-100k-mittens models. The later model improves the accuracy of all three punctuation classes, with the most pronounced improvement for the "period" class (9\% absolute improvement). The improvement for "question mark" and "comma" classes is 6.2\% and 6.8\%, respectively. The improvements stem mostly from the fact that the CNN-100k-mittens predicts punctuation marks missed by the baseline model. In other words, our model inserts punctuation marks in places where the baseline model predicts blanks.

\begin{table}[!htb]
    \caption{Confusion matrix comparison for the CNN-baseline and CNN-100k-mittens models.}
    \label{table:confusion_matrix}
    \begin{minipage}{.49\linewidth}
        % \caption{}
        \newcommand\items{4}   %Number of classes
        \arrayrulecolor{white} %Table line colors
        \noindent\begin{tabular}{cc*{\items}{|E}|}
        \multicolumn{1}{c}{} &\multicolumn{1}{c}{} &\multicolumn{\items}{c}{CNN-baseline} \\ \hhline{~*\items{|-}|}
        \multicolumn{1}{c}{} & 
        \multicolumn{1}{c}{} & 
        \multicolumn{1}{c}{\rot{$\epsilon$}} & 
        \multicolumn{1}{c}{\rot{.}} & 
        \multicolumn{1}{c}{\rot{?}} & 
        \multicolumn{1}{c}{\rot{,}} \\ \hhline{~*\items{|-}|}
        \multirow{\items}{*}{\rotatebox{90}{Actual}} 
        &$\epsilon$ &     95.8 &   1.4 &   0.2 &   2.6 \\ \hhline{~*\items{|-}|}
        & . &     32.2 &  58.7 &   1.5 &   7.6 \\ \hhline{~*\items{|-}|}
        & ? &     23.7 &  21.0 &   49.0 &   6.3 \\ \hhline{~*\items{|-}|}
        & , &     37.8 &   6.5 &   0.6 &  55.0 \\ \hhline{~*\items{|-}|}
        
        \end{tabular}
    \end{minipage}%
    \begin{minipage}{.49\linewidth}
        % \caption{Confusion matrix for the CNN+mittens-100k model.}
        \newcommand\items{4}   %Number of classes
        \arrayrulecolor{white} %Table line colors
        \noindent\begin{tabular}{cc*{\items}{|E}|}
        \multicolumn{1}{c}{} &\multicolumn{1}{c}{} &\multicolumn{\items}{c}{CNN-100k-mittens} \\ \hhline{~*\items{|-}|}
        \multicolumn{1}{c}{} & 
        \multicolumn{1}{c}{} & 
        \multicolumn{1}{c}{\rot{$\epsilon$}} & 
        \multicolumn{1}{c}{\rot{.}} & 
        \multicolumn{1}{c}{\rot{?}} & 
        \multicolumn{1}{c}{\rot{,}} \\ \hhline{~*\items{|-}|}
        \multirow{\items}{*}{\rotatebox{90}{}} 
        & & 95.6   & 1.1     & 0.1 &    2.8 \\ \hhline{~*\items{|-}|}
        & & 21.2     & 67.7    & 1.5 &    9.6   \\ \hhline{~*\items{|-}|}
        & & 19.9      & 18.4      & 55.2 &  6.4   \\ \hhline{~*\items{|-}|}
        & & 31.1     & 6.6     & 0.4  &    61.9   \\ \hhline{~*\items{|-}|}
        \end{tabular}
    \end{minipage} 
\end{table}

\section{Conclusions}\label{sec:conclusions}

Punctuation prediction is an important topic for improving the readability and the segmentation of audio transcripts for downstream processing. Many applications require working with noisy textual data, which poses a~serious challenge to the language models trained on segmented written text. A good example is Automatic Speech Recognition. In addition, ASR introduces additional problems caused by the inherent errors in word recognition and a~significant divergence of punctuation mark distribution due to the dynamics of spontaneous speech. These phenomena limit the usefulness of word embeddings, since most of the static word embeddings are trained on the correctly segmented sentences from written text.

We have hypothesized that aligning language models present in pre-trained word embeddings with the word co-occurrence structure visible in transcribed calls would allow us to overcome some of the challenges of ASR and to improve the quality of punctuation prediction. We have used a~recently published method -- Mittens -- to retrofit the GloVe embeddings with call transcripts obtained from ASR. In this paper, we have shown that the retrofitted embeddings yield an improvement over the original GloVe embeddings. The 6\%-9\% improvement is obtained consistently for all punctuation classes. Furthermore, the retrofitted embeddings allow us to outperform the BiLSTM model with a~faster CNN-100k-mittens model. This is a~very important practical result, as CNN-based models are much more suitable for the deployment in production environments due to the inference time constraints.

Following our findings, we are encouraged to seek further improvements of the established solutions using relatively simple methods prior to migrating to much larger and computationally more expensive models~\cite{merity2019single}. We plan to experiment with retrofitting word embeddings not only on the ASR transcripts, but also all n-best hypotheses produced by ASR. We are also interested in more thorough examination of the effect of retrofitting on downstream tasks other than punctuation prediction, e.g., intent annotation and named entity recognition.

\bibliographystyle{IEEEtran}
\bibliography{punctuation}

\end{document}